\documentclass[letterpaper, 10 pt, conference]{ieeeconf}  

\IEEEoverridecommandlockouts                              

\usepackage{amsmath} 
\usepackage{amsfonts}
\usepackage[dvipsnames]{xcolor}
\usepackage{graphicx}
\usepackage{cuted} 
\usepackage{booktabs}

\title{\LARGE \bf
Semantic and Feature Guided Uncertainty Quantification of Visual Localization for Autonomous Vehicles$^*$
}

\author{Qiyuan Wu$^{1}$ and Mark Campbell$^{1}$
\thanks{$^*$This work was supported by NSF CPS grant 2211599 and NSF FRR grant 2305532. }
\thanks{$^{1}$Qiyuan Wu and Mark Campbell are with Sibley School of Mechanical and Aerospace Engineering, Cornell University.
        {\tt\small \{qw253, mc288\} @cornell.edu}}%
}

\begin{document}

\maketitle
\thispagestyle{empty}
\pagestyle{empty}

\begin{abstract}

The uncertainty quantification of sensor measurements coupled with deep learning networks is crucial for many robotics systems, especially for safety-critical applications such as self-driving cars.
This paper develops an uncertainty quantification approach in the context of visual localization for autonomous driving, where locations are selected based on images. Key to our approach is to learn the measurement uncertainty using light-weight sensor error model, which maps both image feature and semantic information to 2-dimensional error distribution.
Our approach enables uncertainty estimation conditioned on the specific context of the matched image pair, implicitly capturing other critical, unannotated factors (e.g., city vs.\ highway, dynamic vs.\ static scenes, winter vs.\ summer) in a latent manner.
We demonstrate the accuracy of our uncertainty prediction framework using the Ithaca365 dataset, which includes variations in lighting and weather (sunny, night, snowy). Both the uncertainty quantification of the sensor+network is evaluated, along with Bayesian localization filters using unique sensor gating method.
Results show that the measurement error does not follow a Gaussian distribution with poor weather and lighting conditions, and is better predicted by our Gaussian Mixture model.

\end{abstract}


\begin{figure*}[h]
  \centering
  \includegraphics[width=0.8\textwidth]{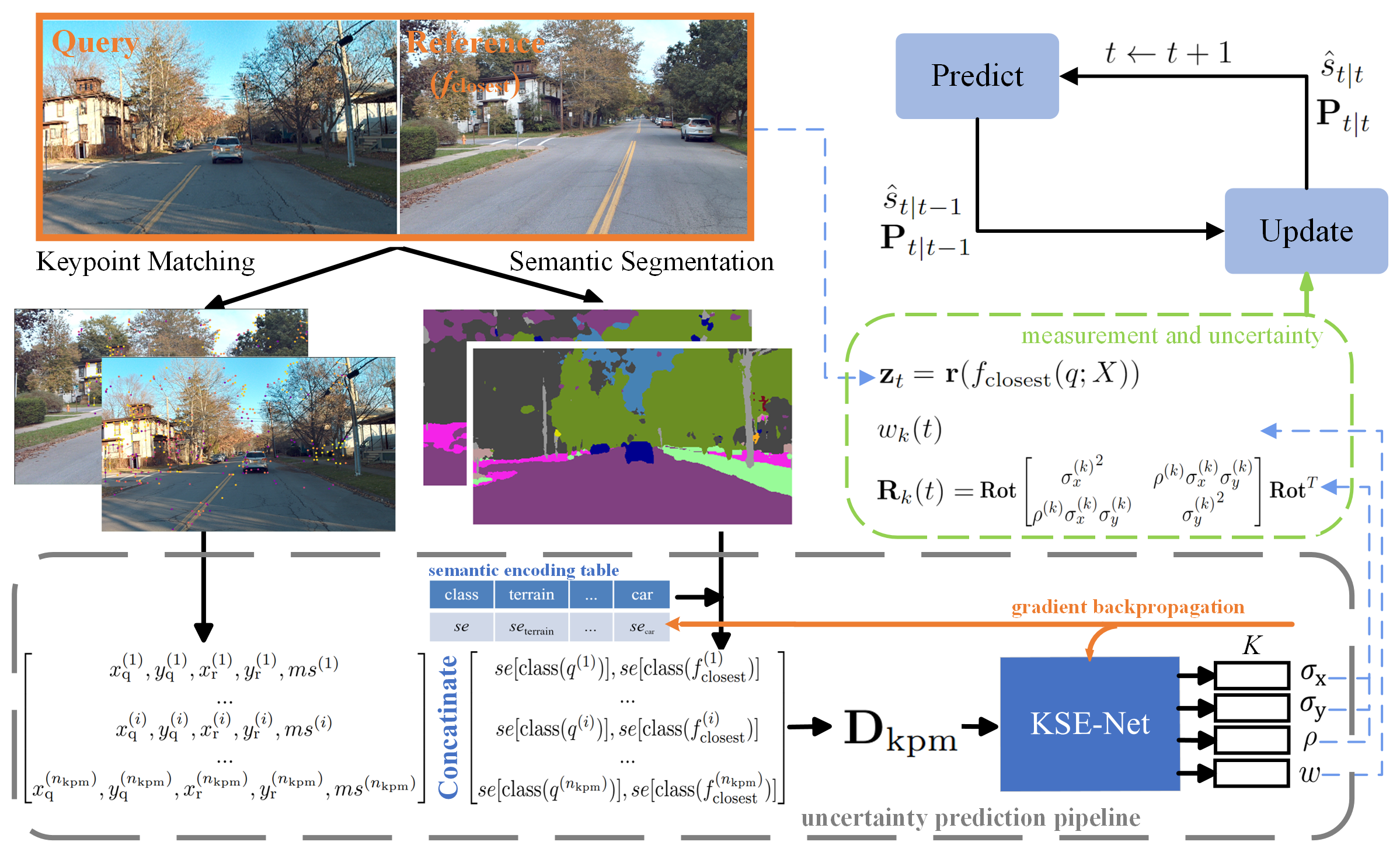}
  \caption{Overall workflow. Input is a pair of matched images (one query image and one closest reference image picked by the deep visual localization pipeline (NetVLAD + Superpoint + SuperGlue)). Keypoint matching is an intermediate output of the mentioned pipeline, and semantic segmentation uses pretrained  DeepLabv3Plus-MobileNet. We develop this uncertainty prediction pipeline in the bottom box, with the semantic encoding table and KSE-Net as learnable parts. During inference, uncertainty is predicted by KSE-Net and measurement is provided by the closest reference image (blue dashed arrows), which is then fed into Bayesian filter for the update step.}
  \label{scheme}
\end{figure*}

\section{Introduction}

 

The evolution of modern large-scale deep learning pipelines has revolutionized the performance of applications ranging from medical diagnostics, business analysis to robotics. However, much of the research in this field has focused on enhancing performance (e.g., average prediction accuracy) through better data collection and architectures.
Despite these advancements, one significant weakness of many large-scale models is their inability to provide a sense of confidence in individual predictions. Predictive accuracies of these models can vary based on factors such as the amount and diversity of training data, the model architecture details, and the complexity of the test environment \cite{wang2020train,hoffman2018cycada}.

In self-driving applications, \textit{probabilistic} uncertainty quantification of deep learning outputs is crucial. Realizing uncertainty models for these networks will not only facilitate their integration into formal probabilistic perception and planning frameworks (e.g., Bayesian filtering) but also enable better safety guarantee and reasoning over the outputs.
While some modern neural networks attempt to output probabilistic uncertainty, the reliability of the uncertainty prediction is still insufficient for safety-critical decision-making \cite{wang2021rethinking}. Most modern neural networks are deterministic or produce only \textit{non-probabilistic} confidence, such as the softmax function.

In this work, we quantify the probabilistic uncertainty of visual localization in the context of autonomous vehicles.
Visual localization is an approach to find the pose of the vehicle using a query (e.g., images or point clouds) which contains key environmental information.
Image-based and 3D-structure-based localization are the two major branches of visual localization; we focus on image-based in this paper.

This paper develops a novel approach to learning model uncertainty for deep visual localization pipelines. Importantly, our approach can be generalized to varying weather and lighting conditions and does not change the original deep visual localization pipeline.
We formulate the problem using a conditional measurement error distribution, similar to the types of  error model specifications provided by manufacturers for traditional sensors, indicating  the accuracy of the sensor under different conditions, e.g., temperature, range, etc. 
In our approach, more specific and complex conditions (e.g., occlusions, scene ambiguity,  lighting) are recognized implicitly by learning from the context within the image frame.
Importantly, our approach to modeling uncertainty can be added on to the network and is thus generalizable to other feature based architectures. We selected the location prediction architecture described in \cite{chen2023probabilistic} as the backbone deep visual localization pipeline. 
A light-weight learning model is used to map both keypoint matches (the output of the original deep visual localization pipeline) and semantic classes (generated from a pretrained model) to predict a 2D non-Gaussian distribution of the measurement error.
We demonstrate the effectiveness of our approach by assessing the accuracy of the model (average and distribution) compared to GPS over varying environmental conditions. We also evaluate the usefulness of the uncertainty model in Bayesian recursive filters - a sigma point filter (SPF) and a Gaussian sum filter (GSF), along with a unique sensor gating method, using a large-scale real-world self-driving dataset with varying weather and lighting conditions.


We present the following contributions of this work:

1) A light-weight, generalizable method for learning  uncertainty models for vision based sensors+networks incorporating both semantic and feature information to predict location uncertainty without modifying the base pipeline;

2) A Gaussian mixture model to most accurately capture measurement uncertainty of the deep visual localization pipeline;

3) Validation of the sensor+network uncertainty model and its incorporation within Bayesian estimation frameworks using a large real-world dataset under various environmental conditions including rain, snow and night settings.


\section{Related Works}


Recent image-based visual localization methods include feature-based, semantic-based, pose regressor and vision transformer methods.
Typical feature-based methods rely on detecting and matching keypoints between images \cite{detone2018superpoint,sarlin2020superglue}, which is simpler and generalizes better to unseen images compared to other methods.
Semantic-based method uses semantically segmented images, together with supplementary information such as semantically labeled 3D point maps of the environment, which is able to achieve accurate localization with less storage space \cite{stenborg2018long}.
Pose regressor learns the pose in an end-to-end (image-to-pose) neural network \cite{kendall2016modelling, zangeneh2023probabilistic}, which can be resource-intensive and unable to deal with dynamic scenes as well as generating to unseen images.
Methods using vision transformer or attention mechanism \cite{wang2024dust3r,giang2023topicfm} do not work well on dynamic scenes either.

Some visual localization approaches consider the uncertainty problem.
\cite{kendall2016modelling} builds an end-to-end Bayesian Neural Network, outputting uncertainty together with prediction.
\cite{zangeneh2023probabilistic} outputs multi-mode distributions for static ambiguous scenes.
But both are limited to static scenes and dense pose annotations.
\cite{xue2023sfd2} gives different reliability weights to different semantic regions, in order to distinguish the uncertainty caused by the moving objects in dynamic scenes.

\section{Method}

Fig.~\ref{scheme} describes the overall workflow of our approach to uncertainty prediction; the image retrieval part (Section \ref{sec:image_retrieval}) is not included for clarity.

\subsection{Location Prediction from Image Retrieval}
\label{sec:image_retrieval}
Let $X = \{I_i\}_{i=1}^N$ be a set of database images with known GPS locations $\mathbf{r}(I_i)$. Given a query image $q$, our goal is to estimate the location where the image was taken. Because images taken from close-by poses should preserve some content similarity, the closest image $f_\mathrm{closest}(q; X)$ is found from database $X$ and its corresponding location is used as the predicted location $\mathbf{\hat{r}}(q) = \mathbf{r}(f_\mathrm{closest}(q; X))$. We define `closest' as the image with the most number of keypoint matches $n_\mathrm{kpm}$ to the query image. To be efficient, global feature matching (NetVLAD \cite{arandjelovic2016netvlad} ) is performed first, followed by neural keypoint matching (SuperPoint \cite{detone2018superpoint} + SuperGlue \cite{sarlin2020superglue} ) on the top $n \ll N$ candidate images. This pipeline is described in detail in \cite{chen2023probabilistic} .

\subsection{Uncertainty Prediction and Quantification}
\label{sec:UQ}

\subsubsection{Problem Definition}

The uncertainty quantification problem is defined as predicting the 2D distribution of location prediction error
\begin{equation}
\label{eq:UQ-define}
    p\left(\mathbf{r}_\mathrm{gt} - \mathbf{\hat{r}} \mid q,f_\mathrm{closest}(q; X)\right)
\end{equation}



\vspace*{0.05 in} 
\subsubsection{Semantics and Pixel Locations of Keypoints}

In \cite{chen2023probabilistic}, number of keypoint matches $n_\mathrm{kpm}$ serves as an indicator for scene similarity and location uncertainty.
However, in certain cases, such as when occlusions are present or in dynamic scenes (e.g., seasonal changes or moving objects),  the number of keypoint matches may provide limited information regarding the uncertainty.
For example, keypoint matches locking on to the front car or repeated ground textures do not provide sufficient information for  location recognition or uncertainty prediction.
In order to mitigate confusion of keypoints on dynamic or ambiguous objects, we use semantic classes of matched keypoints with pixel locations for uncertainty quantification. As shown in Fig.~\ref{scheme}, these are concatenated to form $\mathbf{D}_\mathrm{kpm}(q,f_\mathrm{closest}) $ (Eq.~\eqref{eq:D_kpm}), which is introduced in detail in Section \ref{sec:sensor_error_model}.


\begin{strip}
\vspace*{-0.3 in}
\begin{equation}
\label{eq:D_kpm}
    \mathbf{D}_\mathrm{kpm}(q,f_\mathrm{closest})= 
    \begin{bmatrix}
x_{q}^{(1)} & y_{q}^{(1)} & x_{r}^{(1)} & y_{r}^{(1)} & ms^{(1)} & se[\mathrm{class}(q^{(1)})] & se[\mathrm{class}(f_\mathrm{closest}^{(1)})] \\
    \multicolumn{7}{c}{...} \\
    x_{q}^{(i)} & y_{q}^{(i)} & x_{r}^{(i)} & y_{r}^{(i)} & ms^{(i)} & se[\mathrm{class}(q^{(i)})] & se[\mathrm{class}(f_\mathrm{closest}^{(i)})]  \\
    \multicolumn{7}{c}{\text{...}} \\
    x_{q}^{(n_\mathrm{kpm})} & y_{q}^{(n_\mathrm{kpm})} & x_{r}^{(n_\mathrm{kpm})} & y_{r}^{(n_\mathrm{kpm})} & ms^{(n_\mathrm{kpm})} & se[\mathrm{class}(q^{(n_\mathrm{kpm})})] & se[\mathrm{class}(f_\mathrm{closest}^{(n_\mathrm{kpm})})] 
    \end{bmatrix}
\end{equation}
\vspace*{-0.2 in}
\end{strip}

Pretrained DeepLabv3Plus-MobileNet \cite{chen2018encoder} is used to infer semantic maps, which are then used to assign semantic classes to keypoints.
Instead of one-hot encoding, a scalar value encoding of semantic class is developed in this work. A table of encoding value is learned by the model, creating 19 scalars corresponding to the 19 semantic classes, and the scalar values are associated with keypoints as the representation of semantic classes. By applying this encoding method, the dimension of semantic representation is reduced to 1, instead of 19 with one-hot encoding.
The semantic inferring pipeline is described in the center region of Fig.~\ref{scheme}.

\subsubsection{Sensor error model: KSE-Net}
\label{sec:sensor_error_model}
As in \cite{chen2023probabilistic}, the sensor error can be mapped from the number of keypoint matches, which is an intermediate output of the image retrieval model. But a large amount of data and analysis is required, as a cross-traversal probabilistic analysis must to be performed to create a sensor error model specific to the given traversal. All frames in a single traversal have the same error model, which fails to capture conditions specific to a single frame such as occlusion and scene ambiguity.

In this work, we use a new learning-based sensor error model: Keypoint-Semantic-Error-Net (KSE-Net), which maps pixel locations, matching scores, and the semantic classes of keypoint matches of a single query-reference image pair to a 2D error distribution. The matrix $\mathbf{D}_\mathrm{kpm}(q,f_\mathrm{closest}) \in \mathbb{R}^{n_\mathrm{kpm} \times 7}$ represents these details of keypoint matches and serves as the input of KSE-Net.

Each row of $\mathbf{D}_\mathrm{kpm}(q,f_\mathrm{closest})$ contains information for one keypoint match; all $n_\mathrm{kpm}$ rows form an information matrix for a given image match pair $(q,f_\mathrm{closest})$. The matched keypoints in query ($q$) and reference ($r$) images are denoted as $q^{(\cdot)}$ and $f_\mathrm{closest}^{(\cdot)}$; the superscript ${i}$ indicates the index  of a keypoint match. Here, $x_{\cdot}^{(\cdot)} \in [0, W-1]$ and $y_{\cdot}^{(\cdot)} \in [0, H-1]$ represent  pixel coordinates of matched keypoints,  $ms^{(\cdot)} \in [0,1]$ is the matching score of each keypoint match, and $se[\cdot] \in \mathbb{R}$ is the semantic encoding value, unique to each semantic class.
The images all have a size of $W \times H$ pixels.

   \begin{figure}[h]
      \centering
          \includegraphics[width=0.49\textwidth]{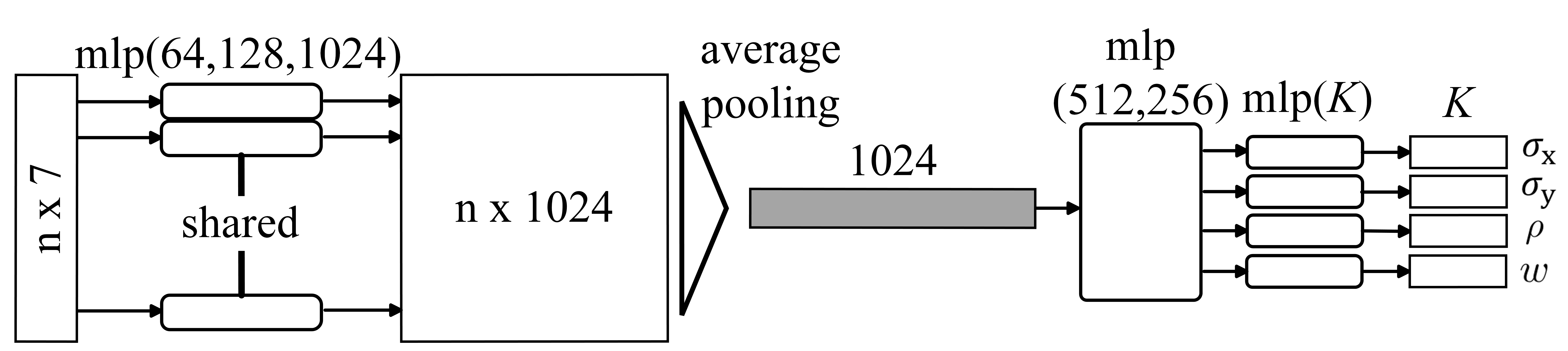}
      \caption{Structure of KSE-Net for uncertainty prediction. } 
      \label{KSE-Net}
   \end{figure}
   
To efficiently handle this type of input, we developed KSE-Net to map keypoint information to a 2D Gaussian mixture distribution of location measurement error, as shown in Fig.~\ref{KSE-Net}; our approach is inspired by PointNet \cite{qi2017pointnet}. The network requires a fixed size input, but the $\mathbf{D}_\mathrm{kpm}$ matrix varies the first dimension based on $N_\mathrm{kpm}$. Thus, cropping, and zero-padding on $\mathbf{D}_\mathrm{kpm}$ is performed after sorting the rows with descending order of $ms^{(\cdot)}$, making the input a fixed size of $Len \times 7$. The outputs of this model are the variances in car lateral direction $\sigma_x^{(k)}$ and  forward direction $\sigma_y^{(k)}$, correlations $\rho^{(k)}$, and weights $w_k$ of each mixture component, forming the error distribution
\begin{equation}
\label{eq:pnetout}
    p(\mathbf{r}_\mathrm{gt}(q) - \mathbf{\hat{r}}(q)) = \sum_{k=1}^{K} w_k \, \mathcal{N}(\mathbf{r}_\mathrm{gt}(q) - \mathbf{\hat{r}}(q) \mid \mathbf{0}, \mathbf{R}_k)
\end{equation}
\begin{equation}
\label{eq:R-GMM}
    \mathbf{R}_k = 
    \begin{bmatrix}
        {\sigma_x^{(k)}}^2 & \rho^{(k)} \sigma_x^{(k)} \sigma_y^{(k)} \\
        \rho^{(k)} \sigma_x^{(k)} \sigma_y^{(k)} & {\sigma_y^{(k)}}^2 \\
    \end{bmatrix}
\end{equation}
With the knowledge of measurement location $\mathbf{\hat{r}}(q) = \mathbf{r}(f_\mathrm{closest}(q; X))$ predicted by the location prediction pipeline (Section \ref{sec:image_retrieval}), the measurement $\mathbf{z}$ can be formulated into a $K$-component 2D Gaussian mixture distribution
\begin{equation}
\label{eq:p_z}
    p(\mathbf{z}) = \sum_{k=1}^{K} w_k \, \mathcal{N}(\mathbf{z} \mid \mathbf{\hat{r}}(q), \mathbf{R}_k)
\end{equation}
which serves as the measurement distribution of the errors.


\subsection{Full Visual Localization Pipeline}

A full visual localization pipeline is built using the location prediction (Section \ref{sec:image_retrieval}) as measurement, the uncertainty prediction (Section \ref{sec:UQ}) as the covariances (measurement uncertainty), within formal estimation frameworks using Sigma Point filter (SPF) (i.e., Unscented Kalman filter (UKF)) \cite{wan2000unscented, brunke2004square} and Gaussian Sum filter (GSF) \cite{alspach1972nonlinear} respectively. The goal of the localization process is to estimate the distribution $p(\mathbf{s}_t|\mathbf{z}_{1:t})$ of the state vector $\mathbf{s}_t$ at time $t$ given observed
measurements $\mathbf{z}_{1:t}$. We define the state vector as:
\begin{equation}
\label{eq:state}
\mathbf{s}=\left[\begin{array}{lllll}
x & y & \theta & v & \dot{\theta}
\end{array}\right]^T
\end{equation}
where $x$, $y$, $\theta$ are the inertial, planar position and heading angle, and $v$, $\dot{\theta}$ are the linear and angular velocity of the car.
In the prediction step of the filter, we assume constant linear and angular velocity ($v$, $\dot{\theta}$) with a small process noise; while better prediction can be realized using IMU measurements, we use this simplified model in order to keep the focus on the vision based measurements and their uncertainties. In the measurement update step, given an image input, we process the image through the location and uncertainty prediction pipeline (Fig.~\ref{scheme}) to give the location measurement $\mathbf{z}=(x, y)+\mathbf{v}$; the uncertainty covariances are transformed from the car coordinate frame to the inertial frame.

\subsubsection{SPF}
The SPF assumes measurement error distribution to be Gaussian, which is simpler and faster compared to GSF.
The output of our uncertainty prediction method is a Gaussian mixture, which can be condensed to a single Gaussian, preserving the first and second moment using the following equations (under condition $\forall i, \boldsymbol{\mu}_i = \mathbf{\hat{r}}(q)$):
\begin{equation}
\label{eq:mu}
    \boldsymbol{\mu} = \mathbf{\hat{r}}(q)
\end{equation}
\begin{equation}
\label{eq:R-SPF}
    \mathbf{R} = \sum_k w_k \mathbf{R}_k
\end{equation}

\subsubsection{GSF}
The GSF is able to process measurements of Gaussian mixture distribution but is more sensitive and involves approximations in the mixture condensation process.

\subsubsection{Gating with Gaussians}

In many real-world scenarios, measurements from sensors may include outliers from noise, unexpected objects, etc.\ that could lead to incorrect state updates. Sensor measurement gating can be applied to the filter to mitigate these effects by using formal hypothesis tests to decide whether to accept a measurement (i.e., use it in the filter update) or reject the measurement (i.e., it is an outlier, outside the nominal error mode).

In the SPF, gating uses the normalized innovations squared $d^2(\mathbf{z}_t)$ as a statistical metric to decide whether to accept/reject a  measurement~\cite{chen2023probabilistic}. This metric is a Chi-square variable; a measurement is rejected if it lies outside the validation gate,
\begin{equation}
\label{eq:gating-cond}
\text { if } d^2(\mathbf{z}_t) >\chi_{\nu, \alpha}^2 \rightarrow \text { reject }
\end{equation}
where $\chi_{\nu, \alpha}^2$ is a threshold from the inverse chi-squared cumulative distribution at a level $\alpha$ with $\nu$ degrees of freedom. Here we take $\nu=2$.
The level $\alpha$ controls the validation gate, i.e. it rejects $(1-\alpha) \times 100 \%$ of the measurements at the tail; typical values are 0.99, 0.975, and 0.95.

\subsubsection{Gating with Gaussian Mixtures (GM gating)}
\label{SEC:GSF-gating}
In the GSF, gating is more complex because there is no analytical solution for the normalized innovations using a Gaussian mixture. We calculate the threshold value $\beta_{\mathrm{thr,}\alpha}$ at a level $\alpha$ numerically by integrating the marginalized distribution $p_\mathbf{d}(x)$ in the direction of the innovation $\mathbf{d} = \frac{\mathbf{z}_t-h(\mathbf{s}_{t \mid t-1})}{\left\|\mathbf{z}_t-h(\mathbf{s}_{t \mid t-1})\right\|}$:
\begin{equation}
\label{eq:p_di}
    p_\mathbf{d}(x) = \sum_{k=1}^{K} w_k \, \mathcal{N}(x \mid \mathbf{d}^T (\boldsymbol{\mu}_k-\bar{\boldsymbol{\mu}}), \mathbf{d}^T \mathbf{\Sigma}_k \mathbf{d})
\end{equation}
\begin{equation}
\label{eq:integration}
    \int_{\beta_{\text{thr,}\alpha}}^{\infty} p_\mathbf{d}(x) \, dx = (1-\alpha)/2
\end{equation}
where $\bar{\boldsymbol{\mu}} = \sum_{k=1}^{K} w_k \boldsymbol{\mu}_k$, and $w_k$ , $\boldsymbol{\mu}_k$ and $\mathbf{\Sigma}_k$ is the weight, mean and covariance of the $k$th mixture component of the original 2D distribution $p(\mathbf{x}) = \sum_{k=1}^{K} w_k \, \mathcal{N}(\mathbf{x} \mid \boldsymbol{\mu}_k, \boldsymbol{\Sigma}_k)$. $K$ is the number of mixtures, and $\mathbf{z}_t$, $h(\cdot)$, and $\mathbf{s}_{t \mid t-1}$ is the measurement, observation function, and state prediction, respectively.
Similar to combining the predicted covariance with the measurement noise covariance ($H \hat{C} H^T+R$) in calculating the normalized innovations squared for SPF gating, we use the combined uncertainty threshold of both prediction and measurement in GM gating. The measurement is rejected if the magnitude of innovation is greater than the sum of the measurement component $\beta_{\text{thr,}\alpha}^{(m)}$ and the predicted component $\beta_{\text{thr,}\alpha}^{(p)}$
\begin{equation}
\label{eq:gating-GSF}
    \text { if } \left\|\mathbf{z}_t-h(\mathbf{s}_{t \mid t-1}) \right\| > \beta_{\text{thr,}\alpha}^{(m)} + \beta_{\text{thr,}\alpha}^{(p)} \rightarrow \text { reject }
\end{equation}
where $\beta_{\text{thr,}\alpha}^{(m)}$ and $\beta_{\text{thr,}\alpha}^{(p)}$ are calculated by Equatino \eqref{eq:integration} on the measurement distribution $p_\mathbf{d}(\mathbf{z})$ and prediction distribution $p_\mathbf{d}(h(\mathbf{s}_{t \mid t-1}))$ respectively.

\section{Experiments}

\subsection{Dataset} 

We use the Ithaca365 dataset \cite{diaz2022ithaca365}, containing data collected over multiple traversals along a 15km route under various conditions: snowy, rainy, sunny, and nighttime. Two types of sensor data are utilized for our experiments: images and GPS locations. Three traversals are randomly selected, with one traversal each from the sunny ($X^1$), nighttime ($X^2$), and snowy ($X^3$), as the reference database. We use three additional traversals ($Q_\text{sunny}^\text{train}$ , $Q_\text{snowy}^\text{train}$ and $Q_\text{night}^\text{train}$), one from each condition, as queries for training and evaluation, and another three traversals ($Q_\text{sunny}^\text{eval}$ , $Q_\text{snowy}^\text{eval}$ and $Q_\text{night}^\text{eval}$) as queries for testing. To avoid double counting and ensure a uniform spatial distribution across the scenes in the evaluation, query images are sampled at an interval of $1$m, except for highways where the spacing is larger. This results in $\approx10,000$ image/GPS pairs for each query traversal.

\begin{figure*}[h!]
  \centering
  \includegraphics[width=\textwidth]{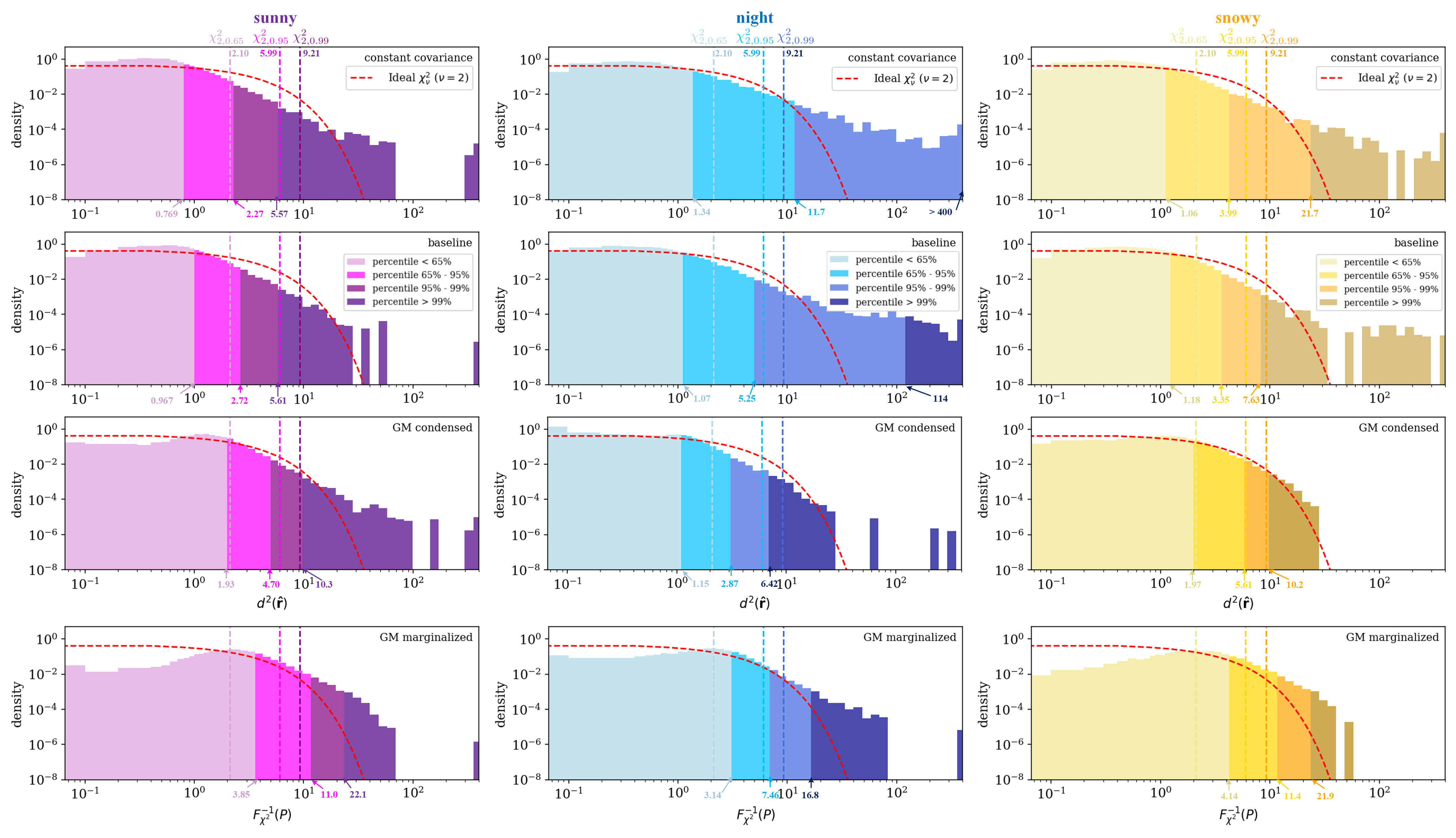}
  \caption{Evaluation of the uncertainty prediction model using the normalized measurement error squared. The histogram bins all measurement instances during a query traversal, normalized to represent a probability density. For constant covariance and baseline, $d^2(\mathbf{\hat{r}})$ is calculated as Equation \eqref{eq:mea_M_dist}. The constant covariance value is obtained by tuning on the validation data.\cite{chen2023probabilistic} For GM condensed, a condensed covariance matrix is calculated as Equation \eqref{eq:R-SPF}. For GM marginalized, the Gaussian mixture distribution is marginalized in the $\mathbf{r}_\mathrm{gt} - \mathbf{\hat{r}}$ direction, and then a confidence percentile ($P$) of ($\mathbf{r}_\mathrm{gt}$) is calculated numerically on the marginalized distribution (1D Gaussian mixture), which finally passes through an inverse Chi-squared function $F_{\chi^2}^{-1}(P)$ to be presented on the horizontal axis. 
  }
  \label{tails}
\end{figure*}

\subsection{Training}

In our uncertainty prediction pipeline, the KSE-Net and the semantic encoding table are trained and evaluated on the dataset of $Q^\text{train} = \{Q_\text{sunny}^\text{train},Q_\text{snowy}^\text{train},Q_\text{night}^\text{train}\}$ as queries and $X = \{X^1,X^2,X^3\}$ as reference database. For each query image $q \in Q^\text{train}$, the closest reference image $f_\mathrm{closest}(q; X)$ is retrieved through the image retrieval pipeline. The information extracted on this pair of matched images $(q,f_\mathrm{closest})$ forms the matrix $\mathbf{D}_\mathrm{kpm}$, which is the input of KSE-Net.

The loss function was negative log-likelihood plus L2 normalization, as shown in
\begin{equation}
\label{eq:lossfunction}
    L = -\log (P_{\mathrm{KSE}(\phi_\mathrm{KSE})}[\mathbf{r}_\mathrm{gt}-\mathbf{\hat{r}} \mid \mathbf{D}_{\mathrm{kpm}, \phi_\mathrm{se}}]) + \lambda_\mathrm{reg} ||\phi_\mathrm{KSE}||^2
\end{equation}
where $\phi_\mathrm{KSE}$ and $\phi_\mathrm{se}$ represents the parameters in KSE-Net and semantic encoding table, respectively, which are the learnable parameters in our model. The KSE-Net output probability $P_{\mathrm{KSE}}$ is described in details in Equation \eqref{eq:pnetout}. $\mathbf{r}_\mathrm{gt}$ is the ground truth location of the query image $q$, and $\lambda_\mathrm{reg}$ is the regularization coefficient.

Adaptive Moment Estimation (ADAM) optimizer was used to optimize the parameters with a learning rate of $10^{-3}$. In our experiments, batch size is set to be 16, $\lambda_\mathrm{reg}$ to be 0.0005, and maximum number of epochs to be 80.

\subsection{Evaluation}

\subsubsection{Uncertainty Prediction Pipeline}
We evaluate our uncertainty prediction model by visualizing the observed distribution of normalized measurement error squared
\begin{equation}
\label{eq:mea_M_dist}
    d^2(\mathbf{\hat{r}}) = (\mathbf{r}_\mathrm{gt} - \mathbf{\hat{r}})^T \mathbf{R}^{-1} (\mathbf{r}_\mathrm{gt} - \mathbf{\hat{r}})
\end{equation}
Fig.~\ref{tails} shows this metric for four different types of uncertainty models (rows: constant covariance, baseline, GM condensed, GM marginalized), and three evaluations (columns: sunny, night, snowy). If the uncertainty quantification is accurate, the density histogram (representing observed probability and confidence percentiles) should ideally be a $\chi_{\nu=2}^2$ distribution (the red dashed curves in Fig.\ref{tails}). The histograms in Fig.~\ref{tails} show that our methods (GM condensed and GM marginalized) produce better probabilistic models of the errors, as evidenced by the smaller distribution tails and more accurate confidence percentiles. In night and snowy conditions, the long tail fails to be captured by the constant and baseline methods, but can be captured by our methods. The GM condensed method predicts much accurate 65\%, 95\%, and 99\% confidence percentiles than the constant covariance and baseline.

\begin{table*}[h!]
\caption{Evaluation result of localization for \textcolor{Purple}{sunny}, \textcolor{Blue}{night}, and \textcolor{Red}{snowy} conditions. Numbers in bold are the best performance among the same gating threshold.}
\label{table_main}
\begin{center}
\resizebox{\textwidth}{!}{%
\begin{tabular}{cccccccccccc}
\toprule
Row & Method & Gating (1 - $\alpha$) & \textcolor{Purple}{$d_\mathrm{err}$(m) $\downarrow$} & \textcolor{Purple}{Cov-credibility(\%)} & \textcolor{Purple}{$n_r$ (\%)} & \textcolor{Blue}{$d_\mathrm{err}$(m) $\downarrow$} & \textcolor{Blue}{Cov-credibility(\%)} & \textcolor{Blue}{$n_r$ (\%)} & \textcolor{Red}{$d_\mathrm{err}$(m) $\downarrow$} & \textcolor{Red}{Cov-credibility(\%)} & \textcolor{Red}{$n_r$ (\%)} \\ 
\midrule
\multicolumn{12}{c}{Average measurement error: \textcolor{Purple}{2718.710 m} / \textcolor{Blue}{2887.717 m} / \textcolor{Red}{2850.384 m}.} \\ \hline
1  & VAPOR \cite{zangeneh2023probabilistic} & 0 & \textcolor{Purple}{2684.698} & \textcolor{Purple}{0.2 / 1.9 / 6.3} & \textcolor{Purple}{0} & \textcolor{Blue}{2876.436} & \textcolor{Blue}{0.2 / 1.1 / 3.9} & \textcolor{Blue}{0} & \textcolor{Red}{2812.506} & \textcolor{Red}{0.2 / 1.2 / 3.9} & \textcolor{Red}{0} \\ 

\hline
\multicolumn{12}{c}{Average measurement error: \textcolor{Purple}{2.734 m} / \textcolor{Blue}{25.596 m} / \textcolor{Red}{5.910 m}.} \\ \hline
2  & baseline \cite{chen2023probabilistic} & 0 & \textcolor{Purple}{0.811} & \textcolor{Purple}{56.4 / 86.3 / 94.0} & \textcolor{Purple}{0} & \textcolor{Blue}{14.219} & \textcolor{Blue}{55.5 / 79.5 / 86.8} & \textcolor{Blue}{0} & \textcolor{Red}{1.752} & \textcolor{Red}{50.0 / 81.3 / 92.1} & \textcolor{Red}{0} \\ 
3  & baseline \cite{chen2023probabilistic} & 1.0\% & \textcolor{Purple}{0.814} & \textcolor{Purple}{57.2 / 87.2 / 94.8} & \textcolor{Purple}{1.6} & \textcolor{Blue}{114.132} & \textcolor{Blue}{48.0 / 66.9 / 72.3} & \textcolor{Blue}{25.2} & \textcolor{Red}{31.632} & \textcolor{Red}{42.3 / 69.9 / 79.6} & \textcolor{Red}{16.5} \\
4  & baseline \cite{chen2023probabilistic} & 2.5\% & \textcolor{Purple}{625.179} & \textcolor{Purple}{41.3 / 62.4 / 67.8} & \textcolor{Purple}{30.4} & \textcolor{Blue}{126.091} & \textcolor{Blue}{47.3 / 65.7 / 71.9} & \textcolor{Blue}{27.2} & \textcolor{Red}{8149.484} & \textcolor{Red}{8.0 / 14.2 / 16.7} & \textcolor{Red}{82.3} \\

\hline
\multicolumn{12}{c}{Average measurement error: \textcolor{Purple}{2.734 m} / \textcolor{Blue}{25.596 m} / \textcolor{Red}{5.910 m}. Number of mixture components $K = 1$} \\ \hline

5  & SPF & 0 & \textcolor{Purple}{0.672} & \textcolor{Purple}{22.2 / 66.8 / 85.9} & \textcolor{Purple}{0} & \textcolor{Blue}{14.247} & \textcolor{Blue}{31.5 / 72.9 / 86.9} & \textcolor{Blue}{0} & \textcolor{Red}{1.493} & \textcolor{Red}{21.9 / 56.7 / 77.8} & \textcolor{Red}{0} \\
6  & GSF & 0 & \textcolor{Purple}{0.672} & \textcolor{Purple}{42.0 / 87.5 / 96.5} & \textcolor{Purple}{0} & \textcolor{Blue}{13.748} & \textcolor{Blue}{51.6 / 86.6 / 92.9} & \textcolor{Blue}{0} & \textcolor{Red}{1.493} & \textcolor{Red}{36.0 / 80.3 / 94.0} & \textcolor{Red}{0} \\
\hline

7  & SPF & 1.0\% & \textcolor{Purple}{0.575} & \textcolor{Purple}{22.3 / 67.2 / 86.3} & \textcolor{Purple}{0.2} & \textcolor{Blue}{1120.036} & \textcolor{Blue}{21.1 / 49.2 / 58.2} & \textcolor{Blue}{37.2} & \textcolor{Red}{\textbf{0.800}} & \textcolor{Red}{22.2 / 57.7 / 78.6} & \textcolor{Red}{0.4} \\
8  & SPF w/ GM gating & 1.0\% & \textcolor{Purple}{\textbf{0.573}} & \textcolor{Purple}{22.4 / 67.2 / 86.3} & \textcolor{Purple}{0.2} & \textcolor{Blue}{25.855} & \textcolor{Blue}{30.9 / 73.0 / 87.4} & \textcolor{Blue}{5.0} & \textcolor{Red}{0.803} & \textcolor{Red}{22.1 / 57.6 / 78.6} & \textcolor{Red}{0.3} \\
9  & GSF & 1.0\% & \textcolor{Purple}{0.612} & \textcolor{Purple}{50.8 / 90.5 / 97.5} & \textcolor{Purple}{14.5} & \textcolor{Blue}{80.074} & \textcolor{Blue}{46.3 / 74.8 / 81.2} & \textcolor{Blue}{21.1} & \textcolor{Red}{0.834} & \textcolor{Red}{38.4 / 82.4 / 95.3} & \textcolor{Red}{6.3} \\
\hline

10  & SPF & 2.5\% & \textcolor{Purple}{0.578} & \textcolor{Purple}{22.3 / 67.2 / 86.3} & \textcolor{Purple}{0.3} & \textcolor{Blue}{221.430} & \textcolor{Blue}{22.9 / 53.1 / 63.2} & \textcolor{Blue}{32.0} & \textcolor{Red}{916.065} & \textcolor{Red}{11.9 / 35.5 / 51.1} & \textcolor{Red}{33.0} \\
11  & SPF w/ GM gating & 2.5\% & \textcolor{Purple}{\textbf{0.576}} & \textcolor{Purple}{22.4 / 67.2 / 86.2} & \textcolor{Purple}{0.2} & \textcolor{Blue}{29.807} & \textcolor{Blue}{30.2 / 71.4 / 84.9} & \textcolor{Blue}{8.3} & \textcolor{Red}{\textbf{0.799}} & \textcolor{Red}{22.2 / 57.7 / 78.6} & \textcolor{Red}{0.4} \\
12  & GSF & 2.5\% & \textcolor{Purple}{58.773} & \textcolor{Purple}{42.9 / 74.0 / 82.9} & \textcolor{Purple}{35.1} & \textcolor{Blue}{47.548} & \textcolor{Blue}{48.8 / 79.2 / 84.9} & \textcolor{Blue}{19.5} & \textcolor{Red}{0.880} & \textcolor{Red}{39.4 / 82.9 / 95.3} & \textcolor{Red}{10.1} \\
\hline

\multicolumn{12}{c}{Average measurement error: \textcolor{Purple}{2.734 m} / \textcolor{Blue}{25.596 m} / \textcolor{Red}{5.910 m}. Number of mixture components $K = 3$} \\ \hline

13  & SPF & 0 & \textcolor{Purple}{\textbf{0.602}} & \textcolor{Purple}{25.8 / 66.7 / 85.8} & \textcolor{Purple}{0} & \textcolor{Blue}{\textbf{2.758}} & \textcolor{Blue}{58.1 / 87.2 / 94.7} & \textcolor{Blue}{0} & \textcolor{Red}{\textbf{0.866}} & \textcolor{Red}{28.8 / 63.9 / 81.8} & \textcolor{Red}{0} \\
14  & GSF & 0 & \textcolor{Purple}{0.727} & \textcolor{Purple}{34.0 / 81.4 / 94.2} & \textcolor{Purple}{0} & \textcolor{Blue}{8.813} & \textcolor{Blue}{50.3 / 89.8 / 97.7} & \textcolor{Blue}{0} & \textcolor{Red}{1.597} & \textcolor{Red}{30.7 / 75.5 / 92.3} & \textcolor{Red}{0} \\
\hline

15  & SPF & 1.0\% & \textcolor{Purple}{0.603} & \textcolor{Purple}{25.8 / 66.7 / 85.8} & \textcolor{Purple}{0.1} & \textcolor{Blue}{3.101} & \textcolor{Blue}{58.4 / 87.1 / 94.5} & \textcolor{Blue}{1.2} & \textcolor{Red}{1.611} & \textcolor{Red}{28.6 / 63.2 / 80.9} & \textcolor{Red}{1.5} \\
16  & SPF w/ GM gating & 1.0\% & \textcolor{Purple}{0.603} & \textcolor{Purple}{25.8 / 66.7 / 85.8} & \textcolor{Purple}{0.1} & \textcolor{Blue}{3.414} & \textcolor{Blue}{57.5 / 86.3 / 93.8} & \textcolor{Blue}{1.4} & \textcolor{Red}{0.865} & \textcolor{Red}{29.0 / 63.9 / 81.8} & \textcolor{Red}{0.2} \\
17  & GSF & 1.0\% & \textcolor{Purple}{0.633} & \textcolor{Purple}{45.6 / 86.4 / 95.7} & \textcolor{Purple}{17.6} & \textcolor{Blue}{\textbf{2.731}} & \textcolor{Blue}{51.3 / 90.4 / 97.7} & \textcolor{Blue}{4.9} & \textcolor{Red}{1.500} & \textcolor{Red}{33.6 / 77.2 / 92.9} & \textcolor{Red}{8.2} \\
\hline

18  & SPF & 2.5\% & \textcolor{Purple}{0.603} & \textcolor{Purple}{25.8 / 66.7 / 85.8} & \textcolor{Purple}{0.1} & \textcolor{Blue}{171.615} & \textcolor{Blue}{35.9 / 56.9 / 62.5} & \textcolor{Blue}{32.2} & \textcolor{Red}{97.361} & \textcolor{Red}{20.3 / 47.2 / 62.1} & \textcolor{Red}{24.4} \\
19  & SPF w/ GM gating & 2.5\% & \textcolor{Purple}{0.603} & \textcolor{Purple}{25.8 / 66.7 / 85.8} & \textcolor{Purple}{0.1} & \textcolor{Blue}{89.007} & \textcolor{Blue}{39.6 / 62.1 / 68.0} & \textcolor{Blue}{24.6} & \textcolor{Red}{1.713} & \textcolor{Red}{28.6 / 63.0 / 80.8} & \textcolor{Red}{1.6} \\
20  & GSF & 2.5\% & \textcolor{Purple}{36.956} & \textcolor{Purple}{41.5 / 75.0 / 84.8} & \textcolor{Purple}{34.3} & \textcolor{Blue}{\textbf{3.092}} & \textcolor{Blue}{51.2 / 90.4 / 97.5} & \textcolor{Blue}{7.9} & \textcolor{Red}{2.639} & \textcolor{Red}{34.5 / 77.2 / 92.2} & \textcolor{Red}{13.7} \\
\bottomrule

\end{tabular}%
}
\end{center}
\end{table*}

\subsubsection{Bayesian Filter}
We evaluate the full visual localization pipeline using the location predictions (Section \ref{sec:image_retrieval}) as measurements in the SPF, GSF, and SPF with GM gating (Section \ref{SEC:GSF-gating}) respectively. 
The localization error $d_\mathrm{err}$ is the average distance error between estimated and ground-truth locations, whereas covariance credibility measures the frequency that the 2D localization error lies within the uncertainty bounds of 68.3\%, 95.4\% and 99.7\% credibility. In the SPF, the bounds are 1-, 2-, and 3-sigma covariance ellipses respectively, while in GSF, the bounds $\beta_{\text{thr,}68.3\%}$,  $\beta_{\text{thr,}95.4\%}$, and  $\beta_{\text{thr,}99.7\%}$ are calculated using Equations \eqref{eq:p_di}, \eqref{eq:integration}.
We also report the frequency of measurement rejection $n_r$.

Table~\ref{table_main} shows the results of our uncertainty models and existing methods. Rows 5-12 create a predicted uncertainty model using the output of KSE-Net as a single Gaussian ($K=1$). Rows 13-20 create a predicted uncertainty model using the output of KSE-Net as a 3-component Gaussian Mixture ($K=3$). Row 1 provides an uncertainty-aware pose regressor \cite{zangeneh2023probabilistic} for comparison, modeling the output samples of the regressor as 2D Gaussian Mixtures ($K=3$) and feeding them into the GSF. We use the three traversals in the reference database to train the pose regressor. The regressor outputs location predictions with huge variances due to scene sparsity and ambiguity. Rows 2-4 provide a baseline for comparison, defined as a 1D statistical sensor error model obtained separately for each traversal using a numerical database of the training data \cite{chen2023probabilistic}.

Each experimental evaluation contains three sets of sensor gating experiments, with probability threshold $1-\alpha$ equals to $0$(non-gating), $1.0\%$, and $2.5\%$ respectively. The SFP and GSF are evaluated in all cases, and SPF with GM gating is evaluated in the cases involving gating. All experiments uses the same measurement locations but different measurement uncertainties.

Analysis of Table \ref{table_main} yields several observations:

First, estimation of night query is significantly improved by our method using a $K=3$ GM uncertainty prediction, both in terms of  localization error $d_\mathrm{err}$ and cov-credibility. The inadequacy to estimate the location and uncertainty of night queries is a major deficiency of the baseline method, which is markedly improve by our method. The improvement of cov-credibility, especially of 99.7\% level, indicates the non-Gaussian nature of the error distribution is better captured with our Gaussian mixture model compared to baseline and single Gaussian model. Although a Gaussian mixture is needed to capture the error distribution, the GSF is not necessarily needed for location estimation. The SPF using the single Gaussian combined from mixture components (Equation \eqref{eq:R-SPF}) works well, even better than GSF in the non-gating case. The GSF works well under 1.0\% and 2.5\% gating.

Second, sensor gating using Gaussian mixture uncertainty prediction models is less sensitive compared to the baseline and single Gaussian models. Only a few cases with $1-\alpha = 2.5\%$ diverge, and other cases maintain the error $d_\mathrm{err}$ roughly the same magnitude as the non-gating case. This indicates better measurement uncertainties predicted by our method.

Third, compared with \cite{chen2023probabilistic}, where 9 traversals/databases are used, similar scale of errors is reached with a database containing only 3 traversals in this work, reducing the data required for online estimation.

Fourth, our uncertainty prediction model generalizes well for different conditions and traversals. The model is trained with a mixture of sunny, night, and snowy queries and can be directly applied to queries for any of these weather conditions. The uncertainty prediction does not require producing specific models for each traversal.











\section{CONCLUSIONS}

In this work, a learning based approach for modeling uncertainty of deep visual localization pipelines is presented, and validated using formal estimation methods (statistical models for the predictions and Bayesian filter estimates) using real-world data.
We develop a learning-based sensor error model: Keypoint-Semantic-Error-Net (KSE-Net), which utilizes both image feature and semantics to predict measurement uncertainty modeled as a 2-dimensional Gaussian mixture distribution.
A large-scale real-world self-driving dataset with varying weather and lighting conditions is used for evaluation. Results demonstrate that the deep learning location predictions are more accurately modeled as a Gaussian mixture, and our approach more accurately captures the longer tails and outliers of the predictions. We also evaluate our approach using a sigma point filter and a Gaussian sum filter, with a unique GM sensor gating method, demonstrating accurate uncertainty predictions across all conditions.
Notably, our approach makes a significant improvement in predicting the location and measurement uncertainty for the night conditions, which is a major deficiency of previous methods. Our approach generalizes well to various conditions, eliminating the onerous work of building traversal-specific sensor error models.

\addtolength{\textheight}{-12cm}   

\bibliographystyle{IEEEtran} 
\bibliography{IEEEabrv,ref}

\begin{thebibliography}{10}
\providecommand{\url}[1]{#1}
\csname url@rmstyle\endcsname
\providecommand{\newblock}{\relax}
\providecommand{\bibinfo}[2]{#2}
\providecommand\BIBentrySTDinterwordspacing{\spaceskip=0pt\relax}
\providecommand\BIBentryALTinterwordstretchfactor{4}
\providecommand\BIBentryALTinterwordspacing{\spaceskip=\fontdimen2\font plus
\BIBentryALTinterwordstretchfactor\fontdimen3\font minus \fontdimen4\font\relax}
\providecommand\BIBforeignlanguage[2]{{%
\expandafter\ifx\csname l@#1\endcsname\relax
\typeout{** WARNING: IEEEtran.bst: No hyphenation pattern has been}%
\typeout{** loaded for the language `#1'. Using the pattern for}%
\typeout{** the default language instead.}%
\else
\language=\csname l@#1\endcsname
\fi
#2}}

\bibitem{wang2020train}
Y.~Wang, X.~Chen, Y.~You, L.~E. Li, B.~Hariharan, M.~Campbell, K.~Q. Weinberger, and W.-L. Chao, ``Train in germany, test in the usa: Making 3d object detectors generalize,'' in \emph{Proceedings of the IEEE/CVF Conference on Computer Vision and Pattern Recognition}, 2020, pp. 11\,713--11\,723.

\bibitem{hoffman2018cycada}
J.~Hoffman, E.~Tzeng, T.~Park, J.-Y. Zhu, P.~Isola, K.~Saenko, A.~Efros, and T.~Darrell, ``Cycada: Cycle-consistent adversarial domain adaptation,'' in \emph{International conference on machine learning}.\hskip 1em plus 0.5em minus 0.4em\relax Pmlr, 2018, pp. 1989--1998.

\bibitem{wang2021rethinking}
D.-B. Wang, L.~Feng, and M.-L. Zhang, ``Rethinking calibration of deep neural networks: Do not be afraid of overconfidence,'' \emph{Advances in Neural Information Processing Systems}, vol.~34, pp. 11\,809--11\,820, 2021.

\bibitem{chen2023probabilistic}
J.~Chen, J.~Monica, W.-L. Chao, and M.~Campbell, ``Probabilistic uncertainty quantification of prediction models with application to visual localization,'' in \emph{2023 IEEE International Conference on Robotics and Automation (ICRA)}.\hskip 1em plus 0.5em minus 0.4em\relax IEEE, 2023, pp. 4178--4184.

\bibitem{detone2018superpoint}
D.~DeTone, T.~Malisiewicz, and A.~Rabinovich, ``Superpoint: Self-supervised interest point detection and description,'' in \emph{Proceedings of the IEEE conference on computer vision and pattern recognition workshops}, 2018, pp. 224--236.

\bibitem{sarlin2020superglue}
P.-E. Sarlin, D.~DeTone, T.~Malisiewicz, and A.~Rabinovich, ``Superglue: Learning feature matching with graph neural networks,'' in \emph{Proceedings of the IEEE/CVF conference on computer vision and pattern recognition}, 2020, pp. 4938--4947.

\bibitem{stenborg2018long}
E.~Stenborg, C.~Toft, and L.~Hammarstrand, ``Long-term visual localization using semantically segmented images,'' in \emph{2018 IEEE international conference on robotics and automation (ICRA)}.\hskip 1em plus 0.5em minus 0.4em\relax IEEE, 2018, pp. 6484--6490.

\bibitem{kendall2016modelling}
A.~Kendall and R.~Cipolla, ``Modelling uncertainty in deep learning for camera relocalization,'' in \emph{2016 IEEE international conference on Robotics and Automation (ICRA)}.\hskip 1em plus 0.5em minus 0.4em\relax IEEE, 2016, pp. 4762--4769.

\bibitem{zangeneh2023probabilistic}
F.~Zangeneh, L.~Bruns, A.~Dekel, A.~Pieropan, and P.~Jensfelt, ``A probabilistic framework for visual localization in ambiguous scenes,'' in \emph{2023 IEEE International Conference on Robotics and Automation (ICRA)}.\hskip 1em plus 0.5em minus 0.4em\relax IEEE, 2023, pp. 3969--3975.

\bibitem{wang2024dust3r}
S.~Wang, V.~Leroy, Y.~Cabon, B.~Chidlovskii, and J.~Revaud, ``Dust3r: Geometric 3d vision made easy,'' in \emph{Proceedings of the IEEE/CVF Conference on Computer Vision and Pattern Recognition}, 2024, pp. 20\,697--20\,709.

\bibitem{giang2023topicfm}
K.~T. Giang, S.~Song, and S.~Jo, ``Topicfm: Robust and interpretable topic-assisted feature matching,'' in \emph{Proceedings of the AAAI conference on artificial intelligence}, vol.~37, no.~2, 2023, pp. 2447--2455.

\bibitem{xue2023sfd2}
F.~Xue, I.~Budvytis, and R.~Cipolla, ``Sfd2: Semantic-guided feature detection and description,'' in \emph{Proceedings of the IEEE/CVF Conference on Computer Vision and Pattern Recognition}, 2023, pp. 5206--5216.

\bibitem{arandjelovic2016netvlad}
R.~Arandjelovic, P.~Gronat, A.~Torii, T.~Pajdla, and J.~Sivic, ``Netvlad: Cnn architecture for weakly supervised place recognition,'' in \emph{Proceedings of the IEEE conference on computer vision and pattern recognition}, 2016, pp. 5297--5307.

\bibitem{chen2018encoder}
L.-C. Chen, Y.~Zhu, G.~Papandreou, F.~Schroff, and H.~Adam, ``Encoder-decoder with atrous separable convolution for semantic image segmentation,'' in \emph{Proceedings of the European conference on computer vision (ECCV)}, 2018, pp. 801--818.

\bibitem{qi2017pointnet}
C.~R. Qi, H.~Su, K.~Mo, and L.~J. Guibas, ``Pointnet: Deep learning on point sets for 3d classification and segmentation,'' 2017.

\bibitem{wan2000unscented}
E.~A. Wan and R.~Van Der~Merwe, ``The unscented kalman filter for nonlinear estimation,'' in \emph{Proceedings of the IEEE 2000 adaptive systems for signal processing, communications, and control symposium (Cat. No. 00EX373)}.\hskip 1em plus 0.5em minus 0.4em\relax Ieee, 2000, pp. 153--158.

\bibitem{brunke2004square}
S.~Brunke and M.~E. Campbell, ``Square root sigma point filtering for real-time, nonlinear estimation,'' \emph{Journal of guidance, control, and dynamics}, vol.~27, no.~2, pp. 314--317, 2004.

\bibitem{alspach1972nonlinear}
D.~Alspach and H.~Sorenson, ``Nonlinear bayesian estimation using gaussian sum approximations,'' \emph{IEEE transactions on automatic control}, vol.~17, no.~4, pp. 439--448, 1972.

\bibitem{diaz2022ithaca365}
C.~A. Diaz-Ruiz, Y.~Xia, Y.~You, J.~Nino, J.~Chen, J.~Monica, X.~Chen, K.~Luo, Y.~Wang, M.~Emond, \emph{et~al.}, ``Ithaca365: Dataset and driving perception under repeated and challenging weather conditions,'' in \emph{Proceedings of the IEEE/CVF Conference on Computer Vision and Pattern Recognition}, 2022, pp. 21\,383--21\,392.

\end{thebibliography}

\end{document}